%% file: main.tex
\documentclass[11pt]{article}
\usepackage{acl}
\input{preamble}

\title{Similar Accuracy, Unequal Evidence: Search APIs as Decision Surfaces for Tool-Using Agents}
\author{Sriram Selvam \and Anneswa Ghosh \\
  Independent Researchers \\
  \texttt{\{selvamsriram,anneswaghosh\}@gmail.com}}
\hypersetup{pdftitle={Similar Accuracy, Unequal Evidence: Search APIs as Decision Surfaces for Tool-Using Agents},pdfauthor={Sriram Selvam and Anneswa Ghosh}}
\newcommand{\ReleaseArtifactLink}{\href{https://github.com/selvamsriram/search-api-decision-surface}{public repository}}
\begin{document}
\maketitle
\input{paper_shared}

\end{document}

%% file: preamble.tex
\usepackage{times}
\usepackage{latexsym}
\usepackage[T1]{fontenc}
\usepackage[utf8]{inputenc}
\usepackage{microtype}
\usepackage{inconsolata}
\usepackage{graphicx}
\usepackage{booktabs}
\usepackage{array}
\usepackage{tabularx}
\usepackage{multirow}
\usepackage{xcolor}
\usepackage{amsmath}
\usepackage{amssymb}
\usepackage{enumitem}
\usepackage{url}
\usepackage{placeins}
\usepackage{pifont}
\usepackage{dblfloatfix}
\usepackage{needspace}
\input{figures/numbers.tex}

\newcolumntype{L}[1]{>{\raggedright\arraybackslash}p{#1}}
\newcolumntype{C}[1]{>{\centering\arraybackslash}p{#1}}
\newcolumntype{R}[1]{>{\raggedleft\arraybackslash}p{#1}}
\newcolumntype{Y}{>{\raggedleft\arraybackslash}X}
\newcolumntype{Z}{>{\raggedright\arraybackslash}X}

\newcommand{\smart}{\textsc{smart}}
\newcommand{\missed}{\textsc{missed}}
\newcommand{\blind}{\textsc{blind}}
\newcommand{\noop}{\textsc{no-op}}
\newcommand{\containsGold}{\texttt{contains\_gold\_answer}}
\newcommand{\contraGold}{\texttt{contradicts\_gold\_answer}}
\newcommand{\goldSnip}{\texttt{gold\_answer\_in\_snippets}}
\newcommand{\goldPage}{\texttt{gold\_answer\_in\_extracted\_page}}
\newcommand{\fetchPage}{\texttt{fetch\_page}}
\newcommand{\searchWeb}{\texttt{search\_web}}
\newcommand{\semanticMatch}{\texttt{semantic\_match}}

\newcommand{\correctans}{\textsc{Correct}}
\newcommand{\tightlist}{}

\makeatletter
\ifacl@linenumbers
  \newenvironment{reviewdisplay}{\begin{nolinenumbers}}{\end{nolinenumbers}}
\else
  \newenvironment{reviewdisplay}{}{}
\fi
\makeatother

\definecolor{AccentBlue}{HTML}{2A80B9}
\definecolor{AccentGreen}{HTML}{2AA876}
\definecolor{AccentOrange}{HTML}{E76F51}

%% file: figures/numbers.tex
\newcommand{\NQueries}{100}

\newcommand{\NTraces}{300}
\newcommand{\NJudgeTotal}{6,909}
\newcommand{\NJudgeValid}{6,869}
\newcommand{\NJudgeInvalid}{40}
\newcommand{\NJudgeSnippetValid}{6,519}
\newcommand{\NJudgePageValid}{350}
\newcommand{\AllCorrect}{10}
\newcommand{\TwoCorrect}{12}
\newcommand{\OneCorrect}{22}
\newcommand{\AllWrong}{56}

\newcommand{\BraveCorrect}{25}

\newcommand{\BraveFone}{0.270}
\newcommand{\BraveSearchAvg}{2.29}
\newcommand{\BraveFetchAvg}{1.02}
\newcommand{\BraveFetchedPct}{65\%}
\newcommand{\BraveAvgTokens}{59,627}
\newcommand{\BraveTokensM}{5.96}
\newcommand{\BraveMedianTokens}{40,638}
\newcommand{\BraveMaxTokens}{303,462}
\newcommand{\BraveOverHundredK}{19}
\newcommand{\BraveFetchSuccess}{92}
\newcommand{\BraveFetchFailed}{10}

\newcommand{\BraveGoldURLExact}{59}
\newcommand{\BraveGoldDomain}{82}
\newcommand{\BraveGoldFamily}{61}
\newcommand{\BraveSnippetSurfaceHit}{71}
\newcommand{\BraveAnswerPage}{52}
\newcommand{\BraveAnswerAvailable}{78}
\newcommand{\BraveWrongWithAnswer}{57}

\newcommand{\BravePreFetchSupportQ}{30}
\newcommand{\BravePostFetchDiscoveredSupportQ}{3}
\newcommand{\BraveTrajectoryVisibleSupportQ}{33}

\newcommand{\BraveSnippetRows}{2,095}
\newcommand{\BravePageRows}{101}
\newcommand{\BraveGoldRows}{97}
\newcommand{\BravePreFetchSupportRows}{101}

\newcommand{\BraveContraRows}{89}

\newcommand{\BraveRankOneGold}{13}
\newcommand{\BraveRankOnePct}{13\%}
\newcommand{\BraveSmartN}{3}
\newcommand{\BraveSmartCorrect}{3}

\newcommand{\BraveSmartRate}{100\%}
\newcommand{\BraveSmartCI}{[44--100]}
\newcommand{\BraveMissedN}{27}
\newcommand{\BraveMissedCorrect}{11}

\newcommand{\BraveMissedRate}{41\%}
\newcommand{\BraveMissedCI}{[25--59]}
\newcommand{\BraveBlindN}{54}
\newcommand{\BraveBlindCorrect}{11}

\newcommand{\BraveBlindRate}{20\%}
\newcommand{\BraveBlindCI}{[12--33]}
\newcommand{\BraveNoopN}{16}
\newcommand{\BraveNoopCorrect}{0}

\newcommand{\BraveNoopRate}{0\%}
\newcommand{\BraveNoopCI}{[0--19]}

\newcommand{\TavilyCorrect}{25}

\newcommand{\TavilyFone}{0.261}
\newcommand{\TavilySearchAvg}{2.74}
\newcommand{\TavilyFetchAvg}{1.30}
\newcommand{\TavilyFetchedPct}{76\%}
\newcommand{\TavilyAvgTokens}{54,156}
\newcommand{\TavilyTokensM}{5.42}
\newcommand{\TavilyMedianTokens}{36,867}
\newcommand{\TavilyMaxTokens}{305,474}
\newcommand{\TavilyOverHundredK}{16}
\newcommand{\TavilyFetchSuccess}{119}
\newcommand{\TavilyFetchFailed}{11}

\newcommand{\TavilyGoldURLExact}{57}
\newcommand{\TavilyGoldDomain}{82}
\newcommand{\TavilyGoldFamily}{60}
\newcommand{\TavilySnippetSurfaceHit}{60}
\newcommand{\TavilyAnswerPage}{57}
\newcommand{\TavilyAnswerAvailable}{75}
\newcommand{\TavilyWrongWithAnswer}{56}

\newcommand{\TavilyPreFetchSupportQ}{16}
\newcommand{\TavilyPostFetchDiscoveredSupportQ}{8}
\newcommand{\TavilyTrajectoryVisibleSupportQ}{24}

\newcommand{\TavilySnippetRows}{2,339}
\newcommand{\TavilyPageRows}{125}
\newcommand{\TavilyGoldRows}{31}
\newcommand{\TavilyPreFetchSupportRows}{34}

\newcommand{\TavilyContraRows}{58}

\newcommand{\TavilyRankOneGold}{17}
\newcommand{\TavilyRankOnePct}{50\%}
\newcommand{\TavilySmartN}{3}
\newcommand{\TavilySmartCorrect}{1}

\newcommand{\TavilySmartRate}{33\%}
\newcommand{\TavilySmartCI}{[6--79]}
\newcommand{\TavilyMissedN}{13}
\newcommand{\TavilyMissedCorrect}{7}

\newcommand{\TavilyMissedRate}{54\%}
\newcommand{\TavilyMissedCI}{[29--77]}
\newcommand{\TavilyBlindN}{67}
\newcommand{\TavilyBlindCorrect}{16}

\newcommand{\TavilyBlindRate}{24\%}
\newcommand{\TavilyBlindCI}{[15--35]}
\newcommand{\TavilyNoopN}{17}
\newcommand{\TavilyNoopCorrect}{1}

\newcommand{\TavilyNoopRate}{6\%}
\newcommand{\TavilyNoopCI}{[1--27]}

\newcommand{\FirecrawlCorrect}{26}

\newcommand{\FirecrawlFone}{0.282}
\newcommand{\FirecrawlSearchAvg}{2.51}
\newcommand{\FirecrawlFetchAvg}{1.28}
\newcommand{\FirecrawlFetchedPct}{81\%}
\newcommand{\FirecrawlAvgTokens}{57,979}
\newcommand{\FirecrawlTokensM}{5.80}
\newcommand{\FirecrawlMedianTokens}{34,383}
\newcommand{\FirecrawlMaxTokens}{380,646}
\newcommand{\FirecrawlOverHundredK}{16}
\newcommand{\FirecrawlFetchSuccess}{121}
\newcommand{\FirecrawlFetchFailed}{7}

\newcommand{\FirecrawlGoldURLExact}{60}
\newcommand{\FirecrawlGoldDomain}{82}
\newcommand{\FirecrawlGoldFamily}{64}
\newcommand{\FirecrawlSnippetSurfaceHit}{54}
\newcommand{\FirecrawlAnswerPage}{61}
\newcommand{\FirecrawlAnswerAvailable}{76}
\newcommand{\FirecrawlWrongWithAnswer}{53}

\newcommand{\FirecrawlPreFetchSupportQ}{16}
\newcommand{\FirecrawlPostFetchDiscoveredSupportQ}{3}
\newcommand{\FirecrawlTrajectoryVisibleSupportQ}{19}

\newcommand{\FirecrawlSnippetRows}{2,085}
\newcommand{\FirecrawlPageRows}{124}
\newcommand{\FirecrawlGoldRows}{27}
\newcommand{\FirecrawlPreFetchSupportRows}{30}

\newcommand{\FirecrawlContraRows}{70}

\newcommand{\FirecrawlRankOneGold}{4}
\newcommand{\FirecrawlRankOnePct}{13\%}
\newcommand{\FirecrawlSmartN}{3}
\newcommand{\FirecrawlSmartCorrect}{1}

\newcommand{\FirecrawlSmartRate}{33\%}
\newcommand{\FirecrawlSmartCI}{[6--79]}
\newcommand{\FirecrawlMissedN}{13}
\newcommand{\FirecrawlMissedCorrect}{7}

\newcommand{\FirecrawlMissedRate}{54\%}
\newcommand{\FirecrawlMissedCI}{[29--77]}
\newcommand{\FirecrawlBlindN}{72}
\newcommand{\FirecrawlBlindCorrect}{18}

\newcommand{\FirecrawlBlindRate}{25\%}
\newcommand{\FirecrawlBlindCI}{[16--36]}
\newcommand{\FirecrawlNoopN}{12}
\newcommand{\FirecrawlNoopCorrect}{0}

\newcommand{\FirecrawlNoopRate}{0\%}
\newcommand{\FirecrawlNoopCI}{[0--24]}

%% file: paper_shared.tex
\begin{abstract}
Search APIs expose ranked snippets, URLs, and metadata on which agents decide whether to answer, search again, or fetch pages. We evaluate these interfaces as \emph{decision surfaces} on a fixed sample of \NQueries{} questions from the 254-question \textsc{SealQA-Hard} subset, using one frozen GPT-5.4 agent, a fixed orchestration harness, and a shared page-fetch backend across Brave, Tavily, and Firecrawl. A Kimi-K2.6 oracle labels visible URL-level evidence; a separate answer audit yields \BraveCorrect{}, \TavilyCorrect{}, and \FirecrawlCorrect{} correct answers out of 100. These counts indicate similar observed accuracy but do not establish equivalence. Under the tested configurations, Brave exposes more pre-fetch support alongside a larger snippet surface; Tavily has a larger rank-1 share among trajectory-pooled supporting observations; and Firecrawl is associated with broader exploration. First-search query-level metrics distinguish support availability from ranking, while contradiction exposure complements contradiction-to-gold ratios. A retrospective oracle union covers 44/100 questions, versus 26/100 for the best individual provider: 18 percentage points of headroom. The observed evidence and action differences motivate evaluating search APIs jointly with agent policy and retrieval budget.
\end{abstract}

\section{Introduction}

Early search-augmented agents often depended primarily on extracted page data provided by search API providers. As complex use cases have evolved, the agent landscape has shifted toward progressive disclosure: models receive top-$n$ URLs, titles, snippets, and key metadata such as freshness, and then decide whether to fetch full pages. In this new landscape, engineering practice often treats search API providers as replaceable components. If two APIs return plausible top-$k$ URLs and produce similar answer accuracy, the rest of the agent pipeline is assumed to be largely unaffected.

This paper argues that the relevant object is the \emph{pre-fetch surface}. Before a page is opened, an agent does not see a search index or a corpus; it relies on search API results. That surface is the evidence state on which the agent decides whether to answer, search again, or spend fetch tokens. We call commercial search APIs \textbf{decision surfaces} for tool-using language agents.

For grounded language-model systems, the retrieval API is part of the grounding interface. It determines which evidence is exposed before generation, which contradictions enter context, and how much retrieval budget is spent before an answer is produced. Decision-surface evaluation therefore measures not only retrieval quality, but also the faithfulness and efficiency conditions under which grounded answers are generated.

The distinction matters because observed correctness can be similar while the internal pipeline diverges. A provider may expose enough snippet evidence for immediate answering. Another may put the relevant URL at rank 1, making a top-result fetch policy effective. A third may expose sparse snippets and be associated with broader exploration under the same policy. These regimes can yield similar aggregate accuracy with different cost, latency, contamination, and failure modes.

Under a controlled protocol, we freeze the answer model, prompt, tools, maximum iterations, judge, and page-fetch backend. The only experimental condition is the commercial search API: Brave, Tavily, or Firecrawl. We analyze URL-level oracle records covering the evidence visible to the agent, separating pre-fetch snippet support from support discovered only after a page fetch. The per-URL oracle lets us ask not only whether the final answer was correct, but also what evidence and actions were available at the decision surface.

This study is a controlled comparison on a fixed sample of \NQueries{} questions from the 254-question \textsc{SealQA-Hard} subset, using one GPT-5.4 agent and a specific orchestration harness. The results characterize the tested provider--harness combinations, rather than establishing a general provider ranking or the best performance achievable with any API. Any of the evaluated APIs may yield higher end-to-end answer accuracy with a different model, prompting strategy, query formulation, search/fetch policy, or retrieval budget. We use the fixed harness to examine how different evidence surfaces interact with the same agent policy.

\paragraph{Contributions.}
\begin{enumerate}[leftmargin=*]
\tightlist
\item We formalize commercial search APIs as \emph{decision surfaces} for tool-using agents rather than static ranked-list retrievers.
\item We introduce a per-URL oracle protocol for auditing visible snippets and fetched pages.
\item We distinguish pre-fetch surface support from post-fetch discovered support, and use the former to define a four-way decision partition (SMART, MISSED, BLIND, NO-OP).
\item We show that similar correctness hides different provider-associated retrieval regimes under a fixed agent: pre-fetch support, rank concentration, blind exploration, contamination, and complementarity.
\end{enumerate}

\section{Related Work}
\label{sec:related}

\paragraph{Retrieval-augmented generation and open-domain QA.}
Retrieval-augmented generation combines parametric models with non-parametric evidence \citep{lewis-etal-2020-retrieval,guu2020realm}. Open-domain QA systems such as DPR, FiD, and Atlas largely evaluate whether a retriever supplies answer-bearing passages to a reader or generator \citep{karpukhin2020dense,izacard-grave-2021-leveraging,izacard2023atlas}. Our setting differs in two ways: retrieval is performed by production web-search APIs, and the reader is an agent that can choose whether to fetch pages.

\paragraph{Information-retrieval evaluation.}
Classical IR metrics, including precision/recall and rank-aware measures such as NDCG, evaluate static rankings \citep{jarvelin2002cumulated,manning2008introduction}. BEIR broadened zero-shot retriever evaluation across tasks and domains \citep{thakur-etal-2021-beir}. Such metrics are necessary but incomplete for agents because they do not ask whether the pre-fetch surface was sufficient, misleading, or action-guiding.

\paragraph{Tool-using and browsing agents.}
WebGPT, Self-Ask, ReAct, IRCoT, Toolformer, and Self-RAG study language models that search, browse, or decide when to call tools \citep{nakano2021webgpt,press2022selfask,yao2023react,trivedi2022ircot,schick-etal-2023-toolformer,asai2023selfrag}. FreshLLMs shows that search augmentation helps with fresh factual knowledge and that evidence order matters \citep{vu2023freshllms}. Over-searching work argues for cost-sensitive evaluation of search-augmented models \citep{xie2026oversearching}. We isolate a complementary variable: the provider surface presented to the same agent.

\paragraph{Hard search benchmarks and evidence evaluation.}
GAIA, WebArena, BrowseComp, and \textsc{SealQA} evaluate agents or systems on tasks requiring search, browsing, or hard factual reasoning \citep{mialon2023gaia,zhou2023webarena,wei2025browsecomp,pham2025sealqa}. RAG evaluation frameworks such as RAGAS and ARES evaluate answer quality, faithfulness, context relevance, or generated claims \citep{es2024ragas,saadfalcon2024ares}. We use a hard-QA benchmark as a controlled probe for provider-associated evidence states under a fixed agent.

\paragraph{Attribution, verifiability, and LLM judges.}
Generative search systems may produce fluent answers with incomplete citation support \citep{liu2023verifiability,yue2023automatic}; long-context work shows that more text does not guarantee robust evidence use \citep{liu2024lost}. LLM-as-judge methods scale evaluation but require constrained rubrics and careful interpretation \citep{zheng2023judging,liu2023geval}. Our judge labels individual retrieved documents with a fixed JSON schema, and our answer correctness label is separately audited as \semanticMatch{}.

\paragraph{Commercial search APIs.}
Brave, Tavily, Firecrawl, and Jina Reader expose different product surfaces \citep{bravesearch-docs,tavily-docs,firecrawl-docs,jina-reader}. We disable provider-side page content in the main condition and use a shared \fetchPage{} backend, so provider differences are attributable to ranked URLs, snippets, and metadata rather than distinct extraction pipelines.

\begin{figure*}[!t]
\centering
\includegraphics[width=0.99\textwidth]{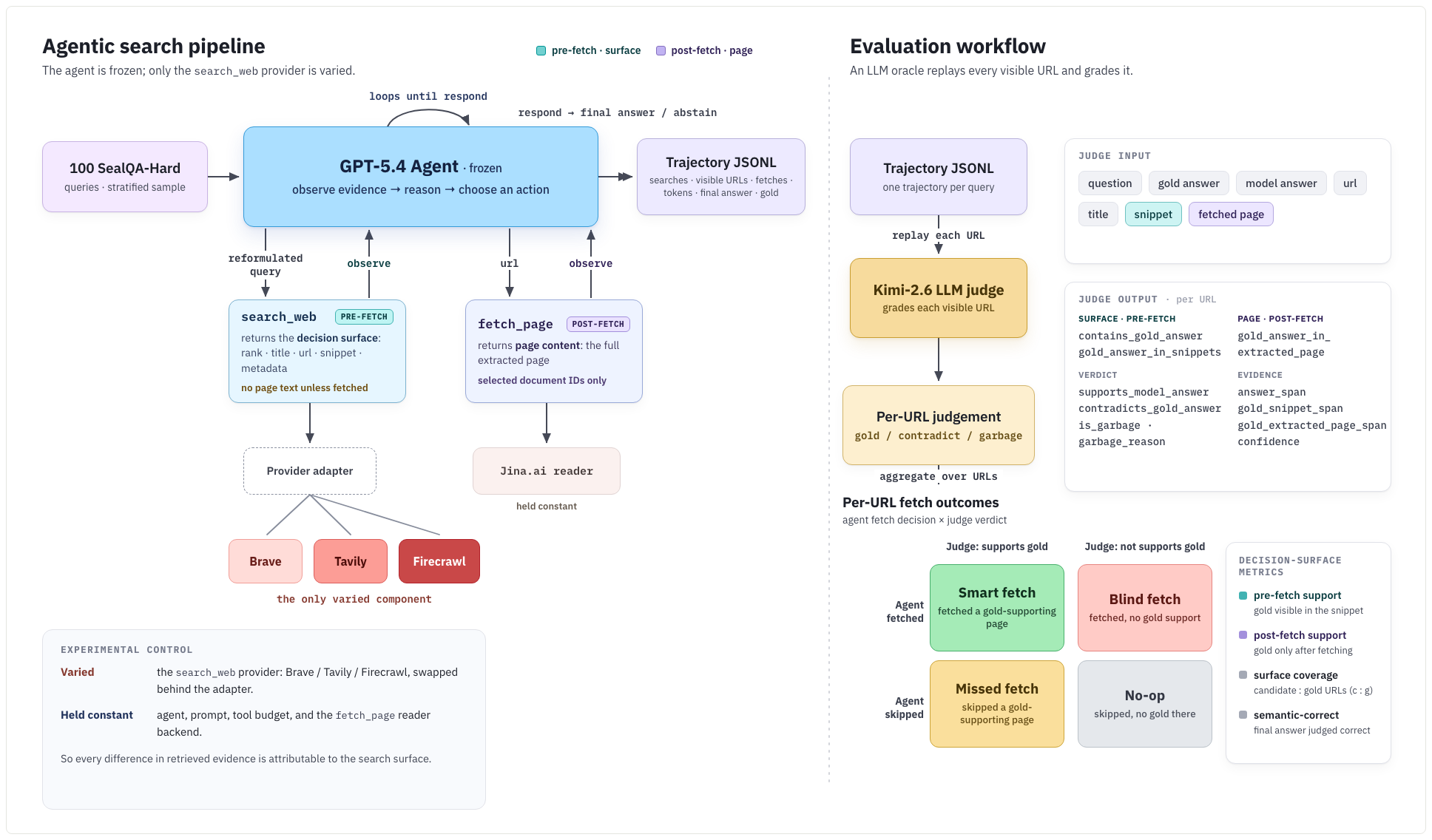}
\caption{Execution and judging pipeline. The agent, prompt, tools, and fetch backend are fixed; only the search provider varies before URL-level records are assembled for oracle analysis.}
\label{fig:architecture}
\end{figure*}

\section{Experimental Protocol}
\label{sec:protocol}

Figure~\ref{fig:architecture} summarizes the experiment. The frozen agent uses the same prompt, tools, and fetch backend in every condition while only the \searchWeb{} provider varies; the analysis pairs URL-level oracle records with logged evidence and actions.

\paragraph{Dataset.}
We use a deterministic proportional stratified sample of \NQueries{} questions from the 254-row \textsc{SealQA-Hard} subset. The strata are \texttt{freshness} $\times$ \texttt{search\_results} $\times$ \texttt{topic}. We chose \textsc{SealQA-Hard} because it provides challenging questions that often require model exploration and are not yet fully resolved by a simple snippet-only search API response. Appendix~\ref{app:dataset} gives the full sampling summary.

\paragraph{Agent and tools.}
The answer model is GPT-5.4 with a maximum of 10 iterations. It has two tools. \searchWeb{} accepts a query and returns up to ten ranked results, each with a document identifier, rank, title, URL, domain, snippet surface, and provider metadata. \fetchPage{} accepts a document identifier from a prior search result and returns extracted markdown plus fetch metadata. The tool interface allows the agent to issue \emph{multiple} \searchWeb{} calls or \emph{multiple} \fetchPage{} calls in a single turn; the trace records each call separately. The final answer must use the exact prefix \texttt{FINAL ANSWER:} and be grounded in retrieved evidence. Appendix~\ref{app:generation} gives the fixed prompt contract and trace schema.

\paragraph{Providers and snippet normalization.}
We compare Brave, Tavily, and Firecrawl Search. Each adapter normalizes provider output into the same result schema. Brave exposes additional provider-native snippet blocks. We aggregate those blocks into the generic \emph{snippet surface} rather than treating them as a separate evidence channel; the agent saw them as snippets, and the judge saw the same text. Thus, we compare configured product surfaces with unequal text volume; snippet normalization does not equalize their evidence budgets. Tavily is run with raw content disabled, and Firecrawl is run without provider-side markdown scraping. Page text visible to the agent therefore comes only from the shared Jina Reader backend.

\paragraph{Run and reproducibility controls.}
The main traces were collected sequentially on 2026-05-17 UTC: Brave 06:49--07:24, Tavily 07:32--08:11, and Firecrawl 08:32--09:02. Runs were not interleaved, so provider and collection-time effects are not fully separated. All providers were asked for up to 10 web results under the fixed configuration in Appendix~\ref{app:providers}: Brave used US/en with moderate safe search, Tavily used basic/general search with raw content disabled, and Firecrawl used web search with provider-side markdown scraping disabled. All full-page text came from the shared Jina Reader \fetchPage{} backend with a 15s timeout, 2MB read limit, concurrency 4, and gzip cache keyed by normalized URL. The release provides code, configuration, schemas, and derived results; complete traces containing provider responses and fetched content are excluded as described in the Reproducibility Statement.

\paragraph{Traces.}
The experiment produces \NTraces{} provider-query trajectories. A trace records every search call, every returned URL, every fetch decision, fetch status, token usage, latency, final answer, and gold answer. Provider-comparison scripts deterministically derive token, URL-hit, and per-query support fields from these traces. Answer correctness comes from the separate semantic answer audit.

\section{Per-URL Oracle and Metrics}
\label{sec:metrics}

\paragraph{Judge input.}
We prepare a Kimi-K2.6 judge record per visible URL. Each judge request contains the question, gold answer, model final answer, and one retrieved document in the same XML-like format the answer model saw. For unfetched URLs, the judge sees title, URL, domain, snippet surface, and metadata. For fetched URLs, the same record also includes extracted markdown; the snippet-specific fields remain labeled separately.

\paragraph{Judge output.}
The judge returns JSON only and is instructed not to use outside knowledge. The required schema is shown below.

\begin{center}
\footnotesize
\begin{tabularx}{\linewidth}{@{}L{0.48\linewidth}Z@{}}
\toprule
\textbf{Field} & \textbf{Meaning} \\
\midrule
\path{contains_gold_answer} & Evidence makes the gold answer answerable for the question. \\
\path{gold_answer_in_snippets} & Provider-returned snippet surface makes the gold answer answerable. \\
\path{gold_answer_in_extracted_page} & Fetched page text makes the gold answer answerable. \\
\path{supports_model_answer} & Evidence supports the agent's final answer. \\
\path{contradicts_gold_answer} & Evidence conflicts with the gold answer. \\
\path{is_garbage} & URL content is unusable or irrelevant. \\
\path{garbage_reason} & Short reason for a garbage label. \\
\path{answer_span} & Text span supporting the model answer. \\
\path{gold_snippet_span} & Snippet span containing gold evidence. \\
\path{gold_extracted_page_span} & Page span containing gold evidence. \\
\path{confidence} & Judge confidence score. \\
\bottomrule
\end{tabularx}
\end{center}

Across providers, \NJudgeValid{} of \NJudgeTotal{} judge rows are valid; \NJudgeInvalid{} invalid rows are excluded. Valid rows are split into \NJudgeSnippetValid{} snippet-only and \NJudgePageValid{} page-visible rows. We use ``gold support'' in the question-conditioned sense: the evidence must make the gold answer answerable for the given question, not merely mention the answer string.

\paragraph{Oracle validation.}
One author manually audited 180 URL-label cases with provider identity hidden, balanced across provider $\times$ surface $\times$ label $\times$ Kimi value (five per leaf slice). Agreement is 164/174 clear cases (94.3\%), with 10 disagreements; six unclear cases (3.3\% of 180) are excluded only from the agreement denominator (Table~\ref{tab:judge-validation}). This is a single-annotator audit of three generic labels, not inter-annotator agreement or direct validation of the two snippet/page-specific support fields. Appendix~\ref{app:examples} gives the audit procedure and scope.

\begin{table}[!t]
\centering
\scriptsize
\setlength{\tabcolsep}{2pt}
\begin{tabularx}{\linewidth}{@{}L{0.40\linewidth}R{0.11\linewidth}R{0.10\linewidth}R{0.10\linewidth}Y@{}}
\toprule
\textbf{Oracle label} & \textbf{Audited} & \textbf{Clear} & \textbf{Agree} & \textbf{Agreement} \\
\midrule
\path{contains_gold_answer} & 60 & 58 & 55 & 94.8\% \\
\path{contradicts_gold_answer} & 60 & 56 & 53 & 94.6\% \\
\path{is_garbage} & 60 & 60 & 56 & 93.3\% \\
\midrule
Overall & 180 & 174 & 164 & 94.3\% \\
\bottomrule
\end{tabularx}
\caption{Human validation of Kimi per-URL oracle labels.}
\label{tab:judge-validation}
\end{table}

\paragraph{Answer metrics.}
The headline \correctans{} metric is \semanticMatch{} from a separate answer audit. Exact matches were accepted automatically; one author reviewed non-exact answers with provider identity hidden. Eleven additional answers received semantic credit (Appendix~\ref{app:examples}). Legacy token F1 and its normalization are reported in Appendix~\ref{app:metrics} and Table~\ref{tab:app-token}.

\paragraph{Support split.}
We separate support by when it became visible. \emph{Pre-fetch surface support} exists for a provider-query pair when a valid judgment marks \goldSnip{} true, or when an unfetched snippet-only judgment marks \containsGold{} true. This captures support in the URL, title, snippet surface, or metadata before the agent decides whether to fetch. \emph{Post-fetch discovered support} exists when no pre-fetch support was present, but a fetched page-visible judgment marks \goldPage{} true. \emph{Trajectory-visible support} is the union of these two events.

\paragraph{Decision partition.}
We count every recorded fetch attempt, including failed fetches and those with invalid judgments. Joining valid support labels to trace actions assigns each provider-query pair to one cell: \smart{} if pre-fetch support exists and the agent fetched a pre-fetch-supporting URL; \missed{} if pre-fetch support exists but the agent fetched none of those URLs; \blind{} if no pre-fetch support exists and the agent fetched at least one URL; and \noop{} if no pre-fetch support exists and no URL was fetched. The partition is descriptive, not causal: a \missed{} query may still be answered correctly from snippets, and a \blind{} query may still succeed if a page reveals support after the fetch.

\paragraph{Surface contradiction-to-gold ratio.}
We define the pooled surface contradiction-to-gold ratio over snippet-only URL observations, counting repeated appearances:
\[
 r_{c:g}=\frac{|\{u:\contraGold(u)\}|}{|\{u:\containsGold(u)\}|}.
\]
The metric depends on the gold answer but is independent of the model's final answer; it uses only pre-fetch surface labels, not page text.

\paragraph{Query-level diagnostics.}
Using pre-fetch labels, we report first-search Support@1, mean reciprocal rank (MRR), and any support over all \NQueries{} questions, with zero for no observed support. Trajectory-level contradiction measures use valid snippet-only observations (definitions and coverage: Appendix~\ref{app:query-diagnostics}).

\section{Results}
\label{sec:results}

\subsection{Similar correctness masks different evidence economies}

\begin{figure*}[!t]
\centering
\includegraphics[width=0.98\textwidth]{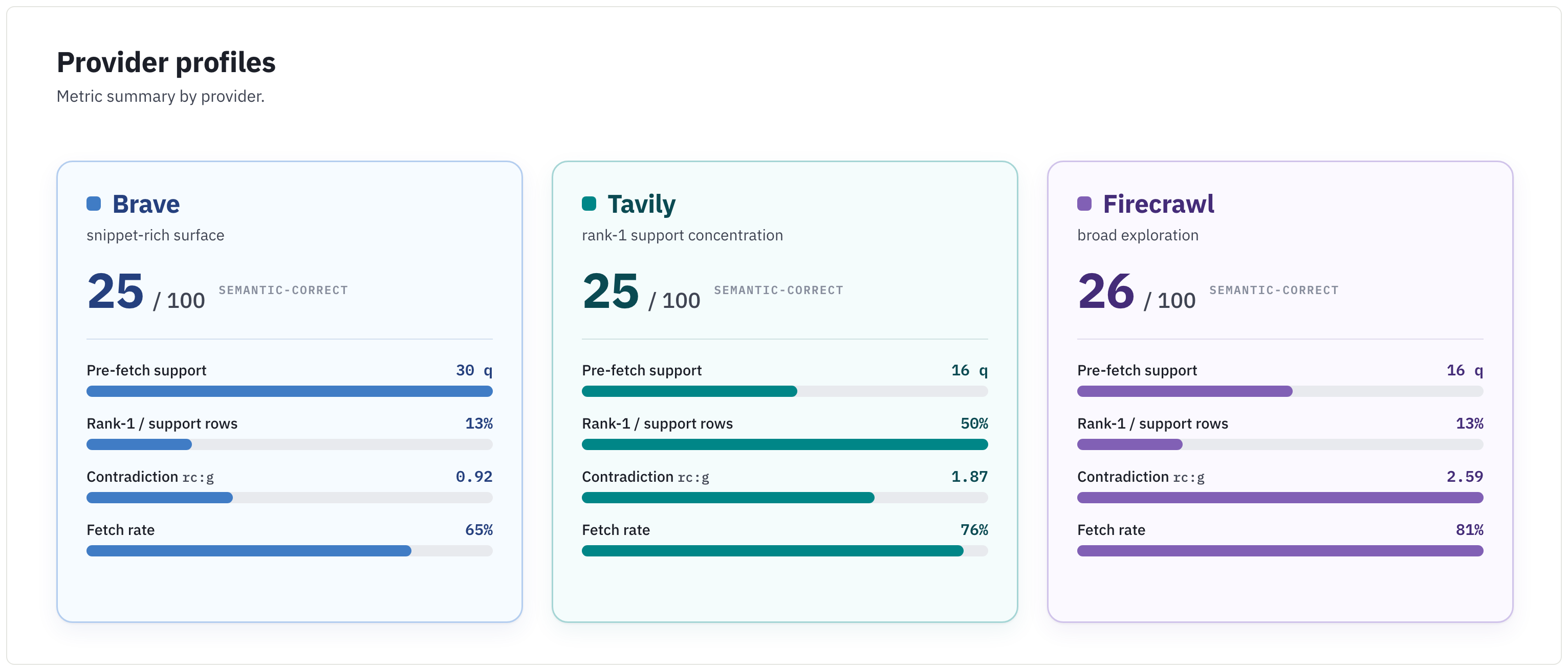}
\caption{Provider profiles under similar semantic correctness. Rank-1 percentages use gold-supporting pre-fetch observations pooled over full trajectories.}
\label{fig:profiles}
\end{figure*}

\begin{table}[!t]
\centering
\scriptsize
\setlength{\tabcolsep}{3pt}
\begin{tabularx}{\linewidth}{@{}lYYY@{}}
\toprule
\textbf{Metric} & \textbf{Brave} & \textbf{Tavily} & \textbf{Firecrawl} \\
\midrule
\correctans{} /100 & \BraveCorrect{} & \TavilyCorrect{} & \FirecrawlCorrect{} \\
Avg. search calls & \BraveSearchAvg{} & \TavilySearchAvg{} & \FirecrawlSearchAvg{} \\
Avg. fetch calls & \BraveFetchAvg{} & \TavilyFetchAvg{} & \FirecrawlFetchAvg{} \\
Fetched queries & \BraveFetchedPct{} & \TavilyFetchedPct{} & \FirecrawlFetchedPct{} \\
Tokens/query & \BraveAvgTokens{} & \TavilyAvgTokens{} & \FirecrawlAvgTokens{} \\
\bottomrule
\end{tabularx}
\caption{Headline metrics. Semantic correctness is similar across providers, while token use, search/fetch behavior, and evidence exposure differ.}
\label{tab:headline}
\end{table}

Table~\ref{tab:headline} and Figure~\ref{fig:profiles} show the headline. Under \correctans{}, the providers remain close: \BraveCorrect{}, \TavilyCorrect{}, and \FirecrawlCorrect{} correct answers out of 100. The low absolute accuracy reflects the intentionally hard \textsc{SealQA-Hard} sample; this setting is useful for our diagnostic purpose because it exposes cases where answer-bearing evidence is visible but not operationally used.

The evidence economies differ. Brave exposes pre-fetch support on \BravePreFetchSupportQ{} queries, compared with \TavilyPreFetchSupportQ{} for Tavily and \FirecrawlPreFetchSupportQ{} for Firecrawl. Across full trajectories, Tavily has the largest rank-1 share among gold-supporting pre-fetch observations (\TavilyRankOnePct{}, versus \BraveRankOnePct{} for Brave and \FirecrawlRankOnePct{} for Firecrawl; counts in Table~\ref{tab:visible-support}). Firecrawl is associated with the largest fetched-query share under this fixed agent policy (\FirecrawlFetchedPct{}). We summarize paired-bootstrap uncertainty for these main contrasts after the support split in Table~\ref{tab:claim-uncertainty}.

\subsection{Pre-fetch support and page-discovered support diverge}

\begin{table}[!t]
\centering
\scriptsize
\setlength{\tabcolsep}{3pt}
\begin{tabularx}{\linewidth}{@{}L{0.34\linewidth}YYY@{}}
\toprule
\textbf{Metric} & \textbf{Brave} & \textbf{Tavily} & \textbf{Firecrawl} \\
\midrule
\multicolumn{4}{@{}l}{\textit{First search call}} \\
Observed support at any rank (\%) & \shortstack[r]{24\\{[16, 33]}} & \shortstack[r]{12\\{[6, 19]}} & \shortstack[r]{9\\{[4, 15]}} \\
Observed Support@1 (\%) & \shortstack[r]{5\\{[1, 10]}} & \shortstack[r]{9\\{[4, 15]}} & \shortstack[r]{2\\{[0, 5]}} \\
Observed MRR & \shortstack[r]{0.103\\{[0.060, 0.154]}} & \shortstack[r]{0.103\\{[0.050, 0.163]}} & \shortstack[r]{0.041\\{[0.014, 0.076]}} \\
\midrule
\multicolumn{4}{@{}l}{\textit{Full trajectory: valid snippet-only observations}} \\
Questions with a contradiction (\%) & \shortstack[r]{27\\{[19, 36]}} & \shortstack[r]{21\\{[13, 29]}} & \shortstack[r]{28\\{[19, 37]}} \\
Contradictory observations (\%) & \shortstack[r]{4.25\\{[2.47, 6.42]}} & \shortstack[r]{2.48\\{[1.28, 3.98]}} & \shortstack[r]{3.36\\{[1.89, 5.17]}} \\
Contradiction/gold $r_{c:g}$ & \shortstack[r]{0.92\\{[0.49, 1.73]}} & \shortstack[r]{1.87\\{[0.83, 4.47]}} & \shortstack[r]{2.59\\{[1.11, 8.00]}} \\
\bottomrule
\end{tabularx}
\caption{First-search ranking and snippet-only contradiction: estimates [95\% question-bootstrap intervals]. Query percentages use all \NQueries{} questions. Coverage: Appendix~\ref{app:query-diagnostics}.}
\label{tab:query-diagnostics}
\end{table}

\begin{table}[!t]
\centering
\footnotesize
\setlength{\tabcolsep}{4pt}
\begin{tabularx}{\linewidth}{@{}L{0.36\linewidth}YYY@{}}
\toprule
\textbf{Metric} & \textbf{Brave} & \textbf{Tavily} & \textbf{Firecrawl} \\
\midrule
Pre-fetch support & \BravePreFetchSupportQ{} & \TavilyPreFetchSupportQ{} & \FirecrawlPreFetchSupportQ{} \\
Post-fetch support & \BravePostFetchDiscoveredSupportQ{} & \TavilyPostFetchDiscoveredSupportQ{} & \FirecrawlPostFetchDiscoveredSupportQ{} \\
Trajectory support & \BraveTrajectoryVisibleSupportQ{} & \TavilyTrajectoryVisibleSupportQ{} & \FirecrawlTrajectoryVisibleSupportQ{} \\
Snippet rows & \BraveSnippetRows{} & \TavilySnippetRows{} & \FirecrawlSnippetRows{} \\
Page-visible rows & \BravePageRows{} & \TavilyPageRows{} & \FirecrawlPageRows{} \\
Rank-1 among gold-supporting pre-fetch rows & \BraveRankOneGold{} / \BravePreFetchSupportRows{} (\BraveRankOnePct{}) & \TavilyRankOneGold{} / \TavilyPreFetchSupportRows{} (\TavilyRankOnePct{}) & \FirecrawlRankOneGold{} / \FirecrawlPreFetchSupportRows{} (\FirecrawlRankOnePct{}) \\
Contradicting rows (snippet-only) & \BraveContraRows{} & \TavilyContraRows{} & \FirecrawlContraRows{} \\
Gold rows (snippet-only) & \BraveGoldRows{} & \TavilyGoldRows{} & \FirecrawlGoldRows{} \\
\bottomrule
\end{tabularx}
\caption{Full-trajectory support and snippet-only counts. Rank-1 percentages are conditional on gold-supporting pre-fetch observations.}
\label{tab:visible-support}
\end{table}

Table~\ref{tab:visible-support} separates the support available before and after fetching: Brave exposes more pre-fetch support, while Tavily recovers more support only after fetch. Table~\ref{tab:query-diagnostics} separates contradiction relative to gold support from question exposure (27\%/21\%/28\%, Brave/Tavily/Firecrawl). Independent-observation deduplication yields ratios of 0.80/1.93/2.50 and the same exposure counts (Appendix~\ref{app:cache-sensitivity}).

First-search diagnostics distinguish availability from position: Brave exposes support on more questions, Tavily has more rank-1 hits, and their observed MRR point estimates are almost identical (Table~\ref{tab:query-diagnostics}).

\begin{table}[!t]
\centering
\scriptsize
\setlength{\tabcolsep}{2pt}
\begin{tabularx}{\linewidth}{@{}L{0.16\linewidth}R{0.13\linewidth}R{0.13\linewidth}R{0.15\linewidth}R{0.16\linewidth}R{0.17\linewidth}@{}}
\toprule
\textbf{Provider} & \textbf{Med. surface tok./q} & \textbf{Mean surface tok./q} & \textbf{Gold-support rows} & \textbf{Gold rows/1K tok.} & \textbf{Support q/1K tok.} \\
\midrule
Brave & 7,810 & 8,968.1 & 101 & 0.113 & 0.033 \\
Tavily & 5,911 & 5,928.4 & 34 & 0.057 & 0.027 \\
Firecrawl & 2,704 & 3,028.1 & 30 & 0.099 & 0.053 \\
\bottomrule
\end{tabularx}
\caption{Snippet-surface length normalization. Token counts are computed over rendered pre-fetch search observations visible to the agent and exclude fetched page text.}
\label{tab:snippet-volume}
\end{table}

The token counts in Table~\ref{tab:snippet-volume} use the rendered pre-fetch \searchWeb{} observations sent to the model, including URLs, titles, domains, snippets, and provider-native extra snippets, while excluding fetched page text and raw provider payloads. Duplicate search observations are counted because they entered the model context. Because Brave exposes more provider-native snippet text, the table normalizes pre-fetch support by rendered snippet-surface token volume. Brave's row-normalized advantage over Tavily persists (0.113 vs. 0.057 gold-supporting rows per 1K tokens), but the contrast with Firecrawl shrinks (0.113 vs. 0.099), and Firecrawl is higher under query-normalized support (0.053 vs. 0.033 support queries per 1K tokens). These descriptive normalizations do not isolate the effect of text volume from retrieval quality; that would require a matched-surface experiment.

\begin{table}[!t]
\centering
\scriptsize
\setlength{\tabcolsep}{2pt}
\begin{tabularx}{\linewidth}{@{}L{0.23\linewidth}L{0.18\linewidth}L{0.12\linewidth}L{0.20\linewidth}Z@{}}
\toprule
\textbf{Claim} & \textbf{Contrast} & \textbf{Diff.} & \textbf{95\% interval} & \textbf{Takeaway} \\
\midrule
Correctness & Brave--Tavily & 0 q & [$-9$, 9] q & includes zero \\
Correctness & Brave--Firecrawl & $-1$ q & [$-10$, 8] q & includes zero \\
Pre-fetch support & Brave--Tavily & +14 q & [4, 24] q & positive \\
Pre-fetch support & Brave--Firecrawl & +14 q & [6, 22] q & positive \\
Rank-1 support share & Tavily--Brave & +37.1 pp & [16.8, 67.3] pp & positive \\
Rank-1 support share & Tavily--Firecrawl & +36.7 pp & [18.8, 66.4] pp & positive \\
\bottomrule
\end{tabularx}
\caption{Paired-bootstrap intervals. Rank-1 shares condition on gold-supporting pre-fetch observations across trajectories.}
\label{tab:claim-uncertainty}
\end{table}

Table~\ref{tab:claim-uncertainty} reports paired-bootstrap intervals over question IDs, with differences shown as left minus right in queries (q) or percentage points (pp). The point estimates are similar, but the paired intervals include zero and also permit differences of several percentage points. They do not establish accuracy equivalence. Intervals remain positive for Brave's pre-fetch-support advantage and Tavily's trajectory-pooled rank-1 share among supporting observations.

\subsection{The same agent enters different decision regimes}

\begin{figure*}[!t]
\centering
\includegraphics[width=0.98\textwidth]{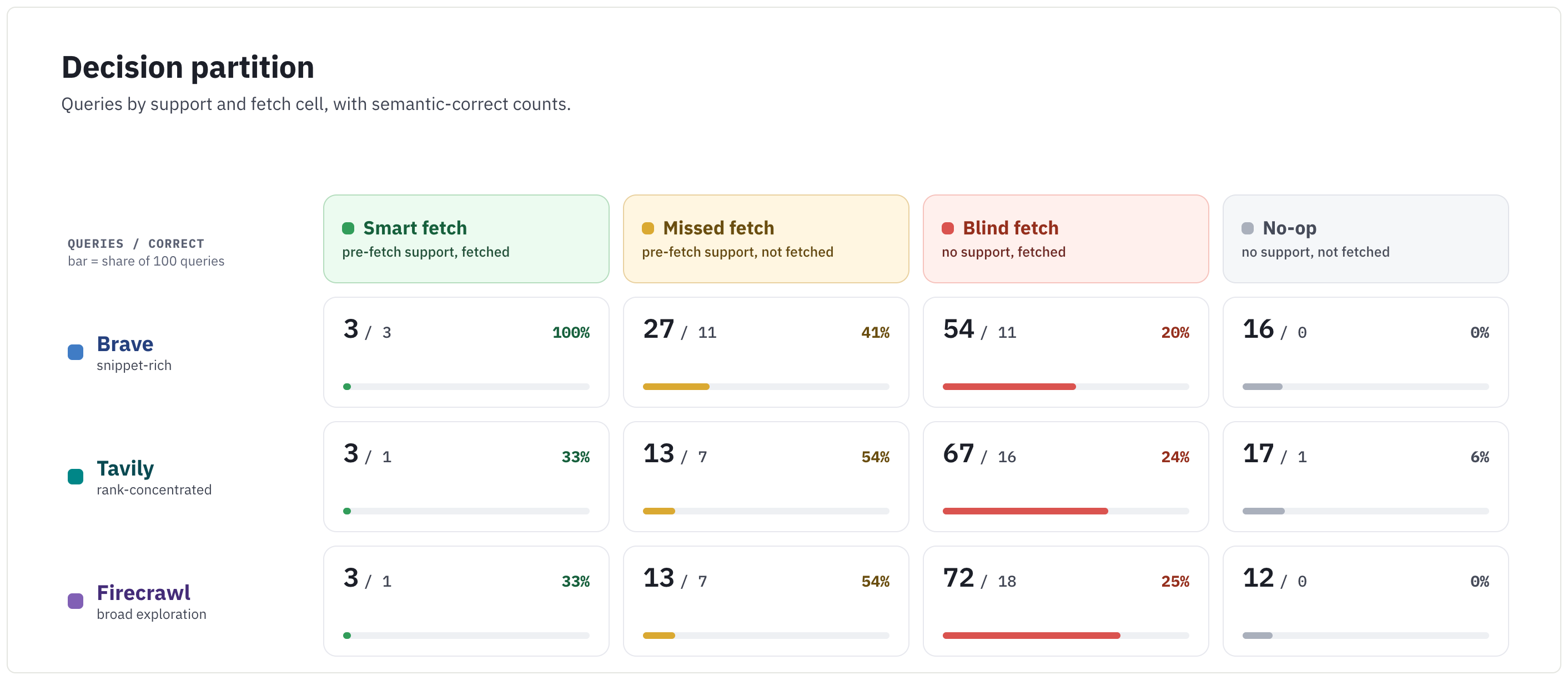}
\caption{Decision partition by provider. Each cell reports queries / semantic-correct answers after joining pre-fetch support labels with recorded fetch attempts, including failed fetches.}
\label{fig:partition}
\end{figure*}

Figure~\ref{fig:partition} describes the observed decision regimes. Each entry is queries / correct answers after joining oracle labels with agent fetch actions. Each provider has only three \smart{} queries, so cell-specific correctness is descriptive and does not establish comparative policy effectiveness. Brave has the largest pre-fetch-support mass, but many such queries are \missed{} rather than \smart{}, consistent with a snippet-rich surface where fetching is not always needed. Firecrawl's largest regime is \blind{} exploration, and Tavily also shifts toward \blind{} once page-discovered support is separated from the pre-fetch surface.

Decision-cell counts are visualized in Figure~\ref{fig:partition}; Wilson intervals are reported in Appendix~\ref{app:additional}, and paired-bootstrap intervals for aggregate metrics are reported in Appendix~\ref{app:uncertainty}.

\subsection{Aggregate correctness hides complementarity}

Only \AllCorrect{} questions are answered correctly by all three providers. \TwoCorrect{} are correct under exactly two providers, \OneCorrect{} under exactly one, and \AllWrong{} under none. The retrospective oracle union is therefore 44/100, compared with 26/100 for the best individual provider: 18 percentage points of headroom. It selects a correct recorded answer using hindsight; it does not demonstrate a deployable router or an accuracy--cost gain. The overlap motivates testing provider routing because provider choice changes \emph{which} questions are solved.

\begin{figure*}[!t]
\centering
\includegraphics[width=0.96\textwidth]{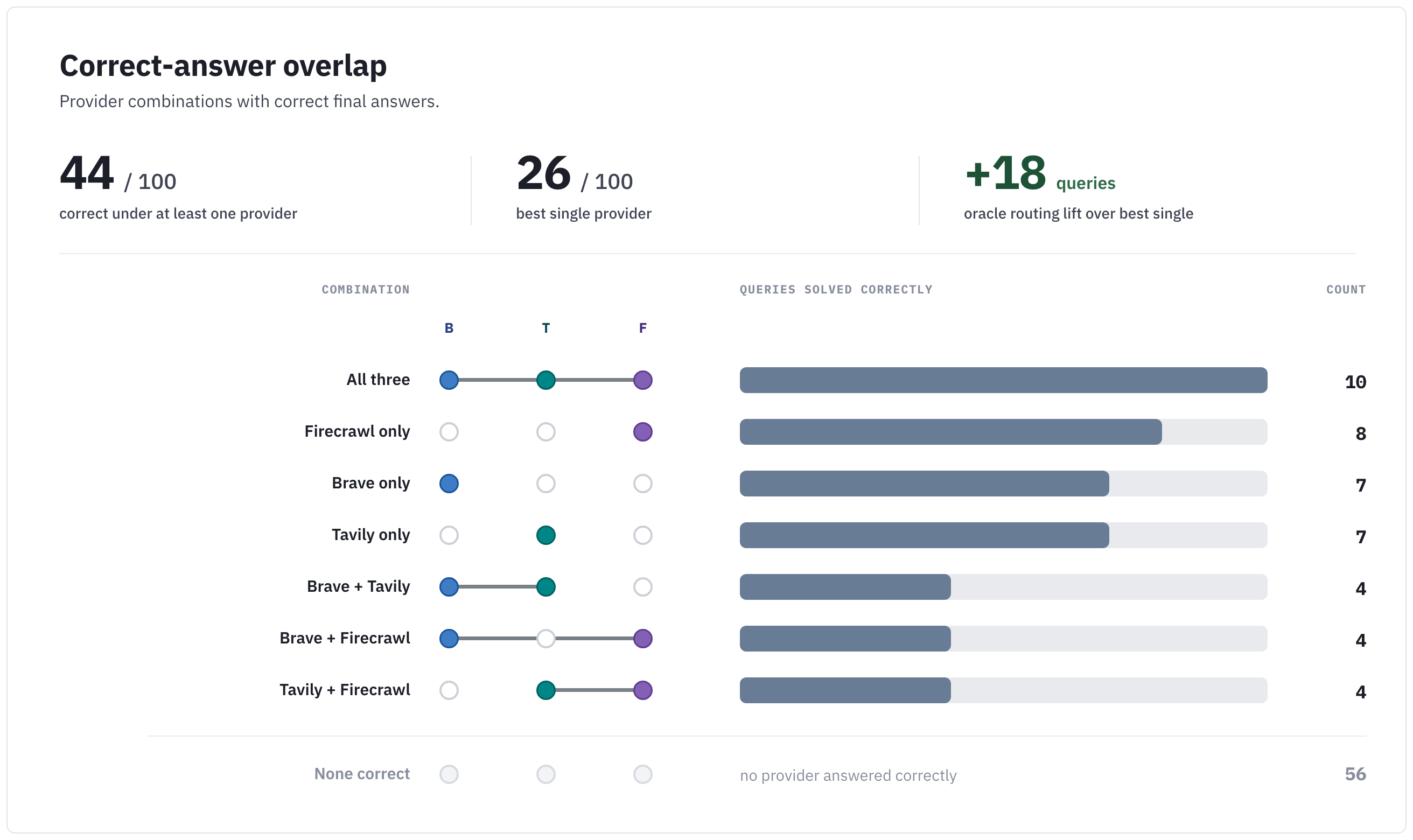}
\caption{Correct-answer overlap across providers. The 44/100 oracle union is a hindsight ceiling over recorded answers, compared with 26/100 for the best individual provider.}
\label{fig:complementarity}
\end{figure*}

\subsection{Many failures occur despite visible answer text}
\label{sec:string-proxy}

The provider-comparison traces include deterministic retrieval proxies such as gold-URL hits and whether the gold answer string appears anywhere in retrieved text. These are \emph{not} the Kimi judge or document-support labels; they are normalized string and URL checks over titles, snippets, provider-returned snippet text, and fetched page text.

Appendix Table~\ref{tab:app-retrieval-diagnostics} shows why ordinary retrieval proxies are incomplete for agents. Gold URLs and answer strings often appear somewhere in the trajectory, yet the agent still has \BraveWrongWithAnswer{}, \TavilyWrongWithAnswer{}, and \FirecrawlWrongWithAnswer{} wrong answers despite visible answer text. String presence alone therefore does not establish usable support or successful evidence use.

\paragraph{Qualitative examples.}
Two Brave traces illustrate competing evidence. In a Blake Shelton query, one snippet listed the five relevant 2015--2023 winning seasons, but nearby snippets repeated his nine lifetime wins; with no fetch, the agent answered \emph{9} rather than \emph{5}. In a border query, rank 1 favored Canada--United States, while lower snippets preserved the key word \emph{continuous} and identified Kazakhstan--Russia; the agent again answered from the generic surface. These cases illustrate relevant evidence co-occurring with distractors; they do not isolate the cause of the final error. Appendix~\ref{app:generation} shows the trace-level artifact structure used for these audits.

\section{Discussion}
\label{sec:discussion}

\paragraph{Provider choice is a policy choice.}
Under the tested policy, Brave exposes more pre-fetch support, Tavily has a larger rank-1 share among trajectory-pooled supporting observations, and Firecrawl is associated with broader exploration. These differences motivate testing provider-aware fetch policies, such as snippet sufficiency checks or rank-based fetch selection. This study does not establish their accuracy--cost benefit.

\paragraph{Top-$k$ relevance is incomplete for agents.}
Static relevance and gold URL hit rates remain useful, but they cannot see whether the surface was already answerable, whether a fetch was worthwhile, or whether contradictory snippets competed with gold support. Decision-surface metrics complement IR metrics by measuring the action state available before the fetch decision.

\paragraph{What providers could optimize.}
An agent-ready search API should optimize not only relevance but also actionability: calibrated snippets, stable document identifiers, source/freshness metadata, low contradiction-to-gold contamination, and rankings that align answer-bearing pages with common agent fetch policies.

\paragraph{What practitioners can use now.}
The metrics here can be computed from traces without rerunning provider APIs. A practitioner can run a small sample through candidate providers, judge visible URLs, and use the resulting surface profile to select policy hypotheses for evaluation on held-out questions.

\section{Conclusion}
Evaluating search APIs for tool-using agents requires looking beyond answer accuracy. On the selected \NQueries{} \textsc{SealQA-Hard} questions under our fixed GPT-5.4 harness and provider configurations, observed correctness counts are close, while pre-fetch support, page-time discovery, rank concentration, contradiction, fetch behavior, and instance-level correctness differ. These findings motivate evaluating search APIs jointly with the models and orchestration policies that consume their evidence surfaces.

\section*{Limitations}
\label{sec:limitations}

\paragraph{Single agent and judge.}
The analysis uses one GPT-5.4 answer model, one orchestration harness, and a Kimi-K2.6 judge. Answer accuracy, evidence exposure, and decision-cell counts depend on these choices. Alternative models, prompts, query strategies, search/fetch policies, or budgets may improve end-to-end performance and change the relative provider profiles. We do not estimate the best achievable performance of any provider.

\paragraph{Lower-bound support.}
Pre-fetch support is measured only over URLs actually returned to the agent, and post-fetch support is a lower bound because unfetched URLs are not page-judged. This prevents us from claiming full provider recall.

\paragraph{Snippet-surface asymmetry.}
Brave returns more provider-native snippet text under our configuration. The length-normalized check in Table~\ref{tab:snippet-volume} is descriptive; without a matched-surface control, we cannot separate the effects of text volume and retrieval quality.

\paragraph{Observational policy analysis.}
The agent's actions are observed, not randomized. We do not intervene on ranks, snippets, or fetch decisions. Causal replay is future work.

\paragraph{Scale and provider coverage.}
The experiment covers a deterministic stratified sample of \NQueries{} questions from the 254-question \textsc{SealQA-Hard} subset, three search APIs, and one collection snapshot. The findings are specific to this sample and experimental configuration and need not generalize to the remaining \textsc{SealQA-Hard} questions, other benchmarks, broader search workloads, or later API versions. Small decision cells are diagnostic; broader coverage and repeated collections would strengthen generalization.

\section*{Ethics Statement}
This work evaluates commercial search APIs and LLM agents on public web-search QA data. We do not claim that one provider is universally better. Provider names are reported because they are experimental conditions and because reproduction requires knowing which APIs produced the surfaces. The study may help reduce unnecessary page fetches and improve handling of contradictory evidence. It could also be misread as a provider leaderboard without respecting the stated limitations; we discourage that use.

\section*{Reproducibility Statement}
The accompanying \ReleaseArtifactLink{} provides experiment code, configurations, prompt templates, the query-sampling procedure, trace and judge schemas, evaluation and bootstrap scripts, figure-generation code, paper sources, and derived results.

To respect provider policies and third-party content rights, we do not redistribute complete execution traces or records containing raw provider responses, search snippets, or fetched page text. This exclusion also applies to copies of that content within judge inputs, audit records, and rendered trajectory views.

The released artifacts document the experimental procedure and reported analyses. Researchers can rerun the protocol using their own API credentials and applicable permissions. Because commercial search APIs are live, time-varying systems, such reruns may produce different retrieved evidence and outcomes.

\bibliography{refs}

\clearpage
\appendix
\setlength{\emergencystretch}{1.5em}
\begin{reviewdisplay}
\input{appendix_shared}
\end{reviewdisplay}

%% file: appendix_shared.tex
\section{Appendix Overview}
\label{app:overview}
The appendix is an audit trail for the main paper rather than a second results narrative. It follows the same order as the experiment: data selection, provider surfaces, response generation, oracle construction, and supporting diagnostics.

\begin{description}[leftmargin=0pt,itemsep=0.25em,topsep=0.25em]
\item[Data and providers.] Appendix~\ref{app:dataset} documents how the \textsc{SealQA-Hard} sample was selected; Appendix~\ref{app:providers} fixes the provider requests and shared page-fetch backend.
\item[Generation and traces.] Appendix~\ref{app:generation} records the agent prompt contract, trace schema, and qualitative audit views.
\item[Metrics and results.] Appendix~\ref{app:metrics} defines the Kimi per-URL oracle, validation checks, and derived metrics; Appendix~\ref{app:additional} collects uncertainty intervals, decision-cell diagnostics, token/fetch summaries, and artifact provenance.
\end{description}

\section{Dataset Construction}
\label{app:dataset}
The experiment uses a deterministic 100-query sample from \textsc{SealQA-Hard}. We assign stable query IDs by hashing question text, form strata over \texttt{freshness} $\times$ \texttt{search\_results} $\times$ \texttt{topic}, allocate slots by largest-remainder proportional rounding, and sample within cells using seed 20260509.

The table below records the concrete sample inputs and audit breadcrumbs so the sampling procedure can be checked without relying on prose alone.

\begin{center}
\footnotesize
\begin{tabularx}{\linewidth}{@{}L{0.34\linewidth}Z@{}}
\toprule
\textbf{Dataset field} & \textbf{Value} \\
\midrule
Dataset & \texttt{vtllms/sealqa} \\
Subset & \texttt{seal\_hard} \\
Source rows & 254 \\
Selected rows & \NQueries{} \\
Seed & 20260509 \\
Primary strata & \texttt{freshness}, \texttt{search\_results}, \texttt{topic} \\
Source file & \path{data/raw/seal-hard.jsonl} \\
Selected sample & \path{data/queries/phase1_100.json} \\
Sampling audit & \begin{tabular}[t]{@{}l@{}}\path{data/100-dataset-selection-}\\\path{rationale.md}\end{tabular} \\
Offline fields & Gold answers, gold URLs, metadata, and source indices are retained for grading and audit only. \\
\bottomrule
\end{tabularx}
\end{center}

The selected sample preserves the source distribution closely. Marginal counts are: 25 fast-changing, 43 slow-changing, and 32 never-changing questions; 57 conflicting and 43 unhelpful search-result labels; topics are Science \& Technology 27, Sports 22, Entertainment 22, Others 12, Politics 9, and History \& Geography 8. Effective years are 61 before 2024, 16 in 2024, and 23 in 2025. Multi-label question-type tags include 72 advanced reasoning, 61 entity/event disambiguation, 13 temporal tracking, 5 cross-lingual reasoning, and 2 false-premise questions.

\section{Provider Settings}
\label{app:providers}
The provider configuration is fixed by \path{config/experiment.yaml} and the provider-specific YAML files under \path{config/providers/}. All providers return web-search results rather than provider-side page extracts, and each adapter normalizes output to the same model-visible schema before rendering.

\paragraph{Brave.}
The adapter issues a GET request to \path{/res/v1/web/search} with count 10, offset 0, US/en locale, \texttt{safesearch=moderate}, \texttt{result\_filter=web}, and text decorations disabled. The rendered snippet surface uses \texttt{description} plus provider-native \texttt{extra\_snippets}.

\paragraph{Tavily.}
The adapter issues a POST request to \path{/search} with \texttt{search\_depth=basic}, \texttt{topic=general}, max 10, answer/images/favicon disabled, and \texttt{include\_raw\_content=false}. The rendered snippet surface uses \texttt{content}.

\paragraph{Firecrawl.}
The adapter issues a POST request to \path{/v2/search} with limit 10, web source only, US region, 60s provider timeout, invalid URLs retained, and provider-side markdown scraping disabled. The rendered snippet surface uses \texttt{description} or metadata description.

The normalized result fields are \texttt{rank}, \texttt{title}, \texttt{url}, \texttt{domain}, \texttt{snippet}, and \texttt{provider\_metadata}. URL normalization removes fragments and common tracking parameters. Brave's additional snippet blocks are rendered as \texttt{<extra\_snippet>} elements because they were visible to the agent as part of the provider surface; we do not reclassify them as fetched-page evidence.

\subsection{Shared Page Fetch Backend}
\label{app:fetch}
All page text in the main experiment comes from the same \fetchPage{} backend, independent of the search provider. The backend is Jina Reader markdown through \path{r.jina.ai}; it is invoked only after the agent selects a prior \texttt{document\_id}. Fetch artifacts are cached as gzip JSON records keyed by normalized URL and backend. Each record stores fetch status, final URL, content type, truncation flag, extractor name, character/token counts, a text hash, latency, and the extracted body. The configured guardrails are a 15s timeout, a 2MB maximum read, and concurrency 4. In the analyzed run, successful fetches are recorded as \texttt{jina\_reader\_markdown}; failures are retained as failed fetches rather than silently dropped.

\section{Response Generation}
\label{app:generation}
\label{app:agent-contract}
The answer model, prompt, tools, and loop budget are fixed across providers. Trace rows preserve the exact rendered request snapshots in \path{iterations[].llm_request.messages}; the prompt templates live under \path{src/searchapi_eval/agent/prompts/}.

The next table lists the fixed generation contract. Its purpose is to show which parts of the agent were held constant while the search provider changed.

\begin{center}
\footnotesize
\begin{tabularx}{\linewidth}{@{}L{0.32\linewidth}Z@{}}
\toprule
\textbf{Component} & \textbf{Fixed value} \\
\midrule
Answer model & GPT-5.4 Azure deployment, temperature 0 \\
Prompt templates & \path{system_fetch_tool.liquid}, \path{user_query.liquid}, \path{search_documents.liquid} \\
Loop budget & 10 iterations, up to 10 results per search call \\
Tools & \searchWeb{} over one configured provider; \fetchPage{} using a model-visible \texttt{document\_id} \\
Turn structure & The model may issue multiple search or fetch calls in one turn; each call is traced separately. \\
Search observation & \texttt{document\_id}, rank, title, URL, domain, snippet surface, and provider metadata \\
Fetch observation & Source provenance, fetch metadata, and extracted markdown/text from the shared backend \\
Final answer rule & Response must begin with \texttt{FINAL ANSWER:} and contain only the short factual answer. \\
Hidden fields & Gold answers and gold URLs are retained in traces for offline grading, never sent to the answer model. \\
\bottomrule
\end{tabularx}
\end{center}

In fetch-tool mode, \searchWeb{} observations are snippet-only. Page text enters the model context only after the model chooses a prior \texttt{document\_id}. This makes fetch behavior observable and keeps provider comparisons from reflecting automatic page inclusion.

\subsection{Trace Schema and Prompt Artifacts}
\label{app:trace-repro}
Each provider-query run is one JSONL trace row using the schema in \path{data/trace-schema-v1.md}. The trace records LLM request snapshots, model responses, tool calls, normalized search responses, raw provider payloads, fetch records, token counts, latency, final answers, and offline gold fields. The model-visible search surface is rendered as \texttt{<search\_documents>} with one \texttt{<document>} block per result. Fetched pages appear as \texttt{<fetched\_page>} blocks with URL provenance, fetch metadata, and extracted body content.

The artifact map below documents the internal trace schema and where information enters the audit pipeline; it does not imply redistribution of the recorded content.

\begin{center}
\footnotesize
\begin{tabularx}{\linewidth}{@{}L{0.36\linewidth}Z@{}}
\toprule
\textbf{Artifact} & \textbf{Where it appears} \\
\midrule
Rendered answer prompt & \path{iterations[].llm_request.messages} \\
Tool schemas & \path{iterations[].llm_request.tools} \\
Normalized search response & \path{retrievals[].search_response.results[]} \\
Raw provider response & \path{retrievals[].search_response.raw_response} in recorded traces \\
Agent fetch decision & \path{fetches[].requested_document_id} and \path{fetches[].reason} \\
Fetched page result & \path{fetches[].page_fetch} and the rendered \texttt{<fetched\_page>} block \\
\bottomrule
\end{tabularx}
\end{center}

The canonical trace files contain 100 Brave rows, 101 Tavily rows, and 100 Firecrawl rows. The extra Tavily row is a recovered transient failure; provider summaries select the latest valid row for each of the 100 query IDs, yielding the \NTraces{} analyzed provider-query trajectories.

\paragraph{Example trajectory.}
One qualitative failure case is \path{sealhard_a2c20fcc3ff1}: ``How many times did Blake Shelton win as a coach on The Voice (U.S.) between 2015 and his departure in 2023?'' The gold answer is 5, while all three provider runs answered 9. In the Brave trace \path{trace_5b6eea4e74d34191a4d275d656f82a18}, the agent used 2 iterations, made 1 search call, made 0 fetch calls, and answered from the pre-fetch surface. The provider-comparison row marks this as wrong despite answer text being visible somewhere in retrieved text. The point of the example is not that a single snippet is decisive, but that the pre-fetch surface mixed a period-specific answer with a tempting lifetime-win distractor.

\subsection{Trace Viewer and Qualitative Audit}
\label{app:trace-viewer}
The rendering script, \path{scripts/render_trace.py}, and local dashboard, \path{scripts/trace_dashboard.py}, support inspection of researcher-generated traces. We used rendered trajectory views to spot-check case studies, verify multi-call turns, and inspect failures where answer text was present but unused. These views are not an independent metric source and are excluded from the release when they reproduce provider content.

\section{Oracle and Metrics}
\label{app:metrics}
\label{app:judge-schema}
The Kimi-K2.6 judge takes one retrieved URL record as input. The exact judge prompt template is \path{src/searchapi_eval/evaluation/prompts/document_support_judge.liquid}; the execution wrapper is \path{scripts/run_llm_judge.py}. The judge receives the question, gold answer, model final answer, and one retrieved document rendered in the same XML-like representation that the answer model saw. Snippet-only records contain title, URL, domain, snippets, extra snippets, and metadata. Page-visible records additionally contain the fetched page block that was returned by \fetchPage{}.

Because the oracle is central to the paper's measurements, the table below states the guardrails used to keep each judge row constrained, parseable, and auditable.

\begin{center}
\footnotesize
\begin{tabularx}{\linewidth}{@{}L{0.34\linewidth}Z@{}}
\toprule
\textbf{Judge guardrail} & \textbf{Implementation} \\
\midrule
No external knowledge & Prompt restricts the judge to supplied document content. \\
Strict output & JSON-only schema with support, contradiction, garbage, spans, and confidence. \\
Valid-row filter & Requires schema version, parsed judgment, provider/query/retrieval/URL IDs, and no execution or parse error. \\
Garbage handling & Deterministic precheck flags failed, unsupported, empty, or media fetches before LLM judging. \\
Judgment reuse & Exact imports and question-URL reuse; see the deduplicated analysis below. \\
Error policy & Invalid judge rows are excluded, not converted into negative evidence. \\
\bottomrule
\end{tabularx}
\end{center}

The output schema fields used in the paper record gold support, snippet support, extracted-page support, model-answer support, contradiction, garbage status, evidence spans, and confidence. Across the main artifacts, \NJudgeValid{} of \NJudgeTotal{} rows are valid, split into \NJudgeSnippetValid{} snippet-only and \NJudgePageValid{} page-visible rows.

\subsection{Deduplicated Contradiction Analysis}
\label{app:cache-sensitivity}
The pooled analysis retains historical judgments, including question-URL reuse that did not require identical snippet text.

To assess repeated-URL weighting, we retain the first valid, independently judged snippet-only observation per provider, question, and normalized URL in canonical retrieval/result order. Exact cache imports qualify only when their full messages match the original request. Selection is independent of label value; all 5,634 retained observations have verified original-call provenance, and every eligible question-URL group is represented.

\begin{center}
\footnotesize
\setlength{\tabcolsep}{3pt}
\begin{tabularx}{\linewidth}{@{}L{0.22\linewidth}R{0.16\linewidth}R{0.16\linewidth}Y@{}}
\toprule
\textbf{Provider} & \textbf{Rows} & \textbf{c/g} & \textbf{Ratio [95\% CI]} \\
\midrule
Brave & 1,723 & 63/79 & 0.80 [0.44, 1.49] \\
Tavily & 2,059 & 54/28 & 1.93 [0.83, 5.00] \\
Firecrawl & 1,852 & 65/26 & 2.50 [1.07, 7.44] \\
\bottomrule
\end{tabularx}
\captionof{table}{Deduplicated contradiction-to-gold sensitivity. Rows are retained snippet-only observations; c/g gives contradiction/gold counts.}
\label{tab:app-deduplicated-ratios}
\end{center}

Intervals use 10,000 paired bootstrap samples of the 100 question IDs (seed 20260628), recomputing the ratio from retained observations in each sample. No replicate has a zero gold denominator. The paired differences are Brave--Tavily $-1.13$ [$-3.90$, $-0.17$], Brave--Firecrawl $-1.70$ [$-6.46$, $-0.44$], and Tavily--Firecrawl $-0.57$ [$-4.06$, $0.85$].

The derivation is implemented in \path{scripts/camera_ready_gate2_audit.py}.

\subsection{Metric Definitions}
For a provider $p$ and query $q$, let $U_{p,q}$ denote the valid judged URLs visible in the trajectory. Unfetched URLs have snippet-only judgments; fetched URLs have page-visible judgments that include both the pre-fetch snippet fields and the fetched page block. Let $G_{\mathrm{pre}}(u)$ be true when the pre-fetch surface supports the gold answer: either \goldSnip{} is true, or a snippet-only judgment marks \containsGold{} true. Let $G_{\mathrm{page}}(u)$ be true when a page-visible judgment marks \goldPage{} true. Let $F(u)$ be true when the trace records a \fetchPage{} attempt for $u$, including failed fetches and URLs without a valid judgment. Support predicates range over valid judged URLs; fetch existence ranges over all trace actions.

For each provider-query pair, pre-fetch support is $\exists u:G_{\mathrm{pre}}(u)$. Post-fetch discovered support is $\neg\exists u:G_{\mathrm{pre}}(u)$ and $\exists u:G_{\mathrm{page}}(u)$. Trajectory-visible support is their union.

The decision-cell table turns these definitions into the four labels used in the main paper. It is descriptive: it classifies what support was visible and what the agent fetched, rather than assigning causal blame.

\begin{center}
\footnotesize
\begin{tabularx}{\linewidth}{@{}L{0.15\linewidth}Z@{}}
\toprule
\textbf{Cell} & \textbf{Condition and interpretation} \\
\midrule
\smart{} & $\exists u:G_{\mathrm{pre}}(u)$ and $\exists u:G_{\mathrm{pre}}(u)\wedge F(u)$: the surface exposed support and the agent attempted to fetch it. \\
\missed{} & $\exists u:G_{\mathrm{pre}}(u)$ and $\neg\exists u:G_{\mathrm{pre}}(u)\wedge F(u)$: pre-fetch support was visible but not fetched. \\
\blind{} & $\neg\exists u:G_{\mathrm{pre}}(u)$ and $\exists u:F(u)$: the agent spent fetch budget without pre-fetch support. \\
\noop{} & $\neg\exists u:G_{\mathrm{pre}}(u)$ and $\neg\exists u:F(u)$: no pre-fetch support and no fetch. \\
\bottomrule
\end{tabularx}
\end{center}

The partition joins normalized URLs after resolving each fetch to its source retrieval and document ID. Fetch actions come directly from the canonical trace, independently of judge validity; invalid judgments remain excluded from support measurements. It is descriptive: a \missed{} query can still be answered correctly from snippets, and a \blind{} query can still succeed if a fetched page reveals evidence not present in snippets. The pooled surface contradiction-to-gold ratio counts snippet-only URL observations, including repeated appearances:
\[
 r_{c:g}=\frac{|\{u:\contraGold(u)\}|}{|\{u:\containsGold(u)\}|}.
\]
It is independent of the model's final answer, but not independent of the gold answer.

For answer-level scoring, exact match and token F1 use the same deterministic normalizer: lowercasing, removing articles and punctuation, and mapping simple number words to digits. Token F1 is the token-overlap F1 between the normalized final answer and normalized gold answer, macro-averaged over the \NQueries{} queries. The semantic \correctans{} metric is separate and comes from the audited \semanticMatch{} labels in \path{results/released/answer_audit.csv}.

\subsection{Human Validation and Semantic Corrections}
\label{app:examples}
The validation queue samples row/label pairs using \path{scripts/build_judge_validation_queue.py}; \path{scripts/judge_validation_app.py} records one human decision per case. One author conducted the review with provider identity hidden. The completed CSV contains 180 cases: five per provider $\times$ surface (snippet-only/page-visible) $\times$ label $\times$ Kimi positive/negative value. The audited labels are \containsGold{}, \contraGold{}, and \texttt{is\_garbage}; \goldSnip{} and \goldPage{} were not directly audited.

The reviewer recorded true, false, or unclear. Six unclear decisions remain in the sample and are excluded only from the agreement denominator. The 174 clear decisions include 164 agreements and 10 disagreements. The table describes this balanced audit sample, not a population-wide error rate or agreement between independent annotators.

\begin{center}
\footnotesize
\begin{tabularx}{\linewidth}{@{}L{0.34\linewidth}R{0.13\linewidth}R{0.13\linewidth}R{0.22\linewidth}@{}}
\toprule
\textbf{Slice} & \textbf{Clear} & \textbf{Agree} & \textbf{Agreement} \\
\midrule
Gold support & 58 & 55 & 94.8\% \\
Contradiction & 56 & 53 & 94.6\% \\
Garbage & 60 & 56 & 93.3\% \\
Overall & 174 & 164 & 94.3\% \\
\bottomrule
\end{tabularx}
\end{center}

For the separate final-answer audit, exact matches were automatically accepted and one author reviewed non-exact answers with provider identity hidden. The audit covers all 300 outcomes: 65 exact matches and 11 additional semantic matches, yielding 25/25/26 correct answers for Brave/Tavily/Firecrawl. Accepted corrections include spacing (\emph{Astra Zeneca}/\emph{AstraZeneca}), entity naming (\emph{UnionPay}/\emph{China UnionPay}), and equivalent numeric expressions (\emph{3}/\emph{3 players}). Recorded notes retain ambiguous specificity (\emph{Punjabi}/\emph{Western Punjabi}), rounded values, hedged ranges, and over-inclusive answers as non-matches. The released CSV preserves the individual labels; no answer labels were changed for the camera-ready version.

\section{Results}
\label{app:additional}
The result appendices report uncertainty, decision-cell diagnostics, token and fetch diagnostics, and artifact provenance. These tables support the main claims without changing the headline conclusions.

\subsection{Query-Level Diagnostics}
\label{app:query-diagnostics}
First-search metrics use the first recorded search call in each canonical provider--question trajectory, before later query reformulation. Support is defined by $G_{\mathrm{pre}}$ in Appendix~\ref{app:metrics}; page-only support does not contribute. Support@1 is the proportion of all 100 questions with observed support at rank 1. Mean reciprocal rank averages $1/r$ for the first observed supporting result, with zero when no support is observed. Support at any rank also uses all 100 questions.

Invalid judgments remain unknown. The table reports their coverage and deterministic lower/upper bounds obtained by completing all missing labels as negative/positive, while holding valid labels fixed. These bounds are separate from the bootstrap intervals. Firecrawl has one empty first-search response, which contributes zero rather than a missing judgment.

\begin{center}
\footnotesize
\setlength{\tabcolsep}{3pt}
\begin{tabularx}{\linewidth}{@{}L{0.42\linewidth}YYY@{}}
\toprule
\textbf{Diagnostic} & \textbf{Brave} & \textbf{Tavily} & \textbf{Firecrawl} \\
\midrule
Valid/returned first-search labels & 995/1000 & 958/969 & 978/990 \\
Unknown rank-1 labels & 0 & 5 & 3 \\
Median first-support rank (supported questions) & 3 ($n=24$) & 1 ($n=12$) & 3 ($n=9$) \\
\midrule
Support@1 bounds (\%) & 5--5 & 9--14 & 2--5 \\
Mean reciprocal rank bounds & 0.103--0.107 & 0.103--0.157 & 0.041--0.079 \\
Support-at-any-rank bounds (\%) & 24--26 & 12--19 & 9--17 \\
Contradiction-exposure bounds (\%) & 27--32 & 21--26 & 28--36 \\
\bottomrule
\end{tabularx}
\captionof{table}{First-search coverage, conditional first-support rank, and missing-label bounds. Contradiction exposure uses snippet-only observations over the full trajectory. Bounds concern missing judgments; they are not confidence intervals.}
\label{tab:app-query-coverage}
\end{center}

Contradiction diagnostics use valid snippet-only observations over the full trajectory, counting repeated appearances. Page-visible contradiction labels may reflect page content, so those observations are excluded from this snippet-only analysis. Every question has at least one valid snippet-only observation. Brave/Tavily/Firecrawl have 89/58/70 contradictory observations out of 2,095/2,339/2,085 valid observations, with 97/31/27 gold-supporting observations and 6/12/14 invalid observations. The independent-observation selection in Appendix~\ref{app:cache-sensitivity} leaves question-exposure counts at 27/21/28; the corresponding contradictory-observation percentages are 3.66/2.62/3.51.

Intervals use 10,000 matched-question percentile-bootstrap samples, seed 20260628, recomputing ratios from observation counts in each sample. For Brave--Tavily, Brave--Firecrawl, and Tavily--Firecrawl, respectively, Support@1 differences are $-4$ [$-11$, 3], 3 [0, 7], and 7 [2, 13] percentage points; MRR differences are $-0.0004$ [$-0.067$, 0.064], 0.062 [0.023, 0.107], and 0.063 [0.008, 0.124]; contradiction-exposure differences are 6 [$-3$, 15], $-1$ [$-10$, 8], and $-7$ [$-15$, 1] percentage points. These intervals describe the observed metrics; the missing-label bounds above address incomplete judgments separately.

The first-search query strings are identical across all three providers for 74 questions. Within this subset, observed Support@1 is 4/74, 8/74, and 2/74, support at any rank is 20/74, 10/74, and 7/74, and MRR is 0.117, 0.119, and 0.047 (Brave/Tavily/Firecrawl). This descriptive check retains the distinction between support availability and position. Per-question results, the full first-support-rank distribution, and paired intervals from \path{scripts/camera_ready_gate3_diagnostics.py} are released under \path{results/released/}.

\subsection{Paired Bootstrap Uncertainty}
\label{app:uncertainty}
We resample question IDs with replacement, preserving the matched Brave/Tavily/Firecrawl rows for each question. The tables use 10,000 percentile-bootstrap replicates with seed 20260628. Count metrics are reported per 100 sampled questions; pairwise differences are left minus right in the row's native units.

The first uncertainty table gives provider-level intervals for the main quantities. These intervals show the scale of each provider's measured regime before any pairwise comparison is made. The rank-1 row uses the same denominator as Table~\ref{tab:visible-support}: gold-supporting pre-fetch rows.

\begin{center}
\footnotesize
\setlength{\tabcolsep}{3pt}
\begin{tabularx}{\linewidth}{@{}L{0.34\linewidth}Z@{}}
\toprule
\textbf{Metric} & \textbf{Provider intervals} \\
\midrule
\correctans{} /100 & B 25 [17, 34]; T 25 [17, 34]; F 26 [18, 35] \\
Pre-fetch support /100 & B 30 [21, 39]; T 16 [9, 23]; F 16 [9, 23] \\
Rank-1 among pre-fetch support rows & B 12.9 [7.1, 19.4]; T 50.0 [31.2, 78.9]; F 13.3 [4.2, 26.3] \\
$r_{c:g}$ & B 0.92 [0.49, 1.73]; T 1.87 [0.83, 4.47]; F 2.59 [1.11, 8.00] \\
Fetched queries /100 & B 65 [56, 74]; T 76 [67, 84]; F 81 [73, 89] \\
Avg. fetch calls & B 1.02 [0.81, 1.25]; T 1.30 [1.06, 1.55]; F 1.28 [1.08, 1.50] \\
Tokens/query & B 59,627 [47,706, 72,908]; T 54,156 [44,304, 64,991]; F 57,979 [44,635, 72,878] \\
\bottomrule
\end{tabularx}
\captionof{table}{Provider bootstrap 95\% intervals. B=Brave, T=Tavily, F=Firecrawl.}
\label{tab:app-bootstrap-provider}
\end{center}

The second uncertainty table reports matched pairwise differences. This comparison clarifies the main claim boundary: intervals for correctness differences include zero and do not establish equivalence, while evidence-economy contrasts are more stable.

\begin{center}
\footnotesize
\setlength{\tabcolsep}{3pt}
\begin{tabularx}{\linewidth}{@{}L{0.34\linewidth}Z@{}}
\toprule
\textbf{Metric} & \textbf{Pairwise differences} \\
\midrule
\correctans{} /100 & B--T 0 [$-9$, 9]; B--F $-1$ [$-10$, 8]; T--F $-1$ [$-10$, 8] \\
Pre-fetch support /100 & B--T 14 [4, 24]; B--F 14 [6, 22]; T--F 0 [$-8$, 8] \\
Rank-1 among pre-fetch support rows & B--T $-37.1$ [$-67.3$, $-16.8$]; B--F $-0.5$ [$-13.3$, 10.1]; T--F 36.7 [18.8, 66.4] \\
$r_{c:g}$ & B--T $-0.95$ [$-3.22$, 0.00]; B--F $-1.68$ [$-6.78$, $-0.36$]; T--F $-0.72$ [$-4.86$, 0.53] \\
Fetched queries /100 & B--T $-11$ [$-19$, $-3$]; B--F $-16$ [$-25$, $-7$]; T--F $-5$ [$-12$, 1] \\
Avg. fetch calls & B--T $-0.28$ [$-0.55$, $-0.02$]; B--F $-0.26$ [$-0.51$, $-0.02$]; T--F 0.02 [$-0.23$, 0.27] \\
Tokens/query & B--T 5,471 [$-4{,}591$, 16,176]; B--F 1,648 [$-11{,}834$, 13,959]; T--F $-3{,}823$ [$-18{,}019$, 9,064] \\
\bottomrule
\end{tabularx}
\captionof{table}{Paired bootstrap differences; left minus right.}
\label{tab:app-bootstrap-diff}
\end{center}

\subsection{Additional Diagnostics and Artifact Manifest}
The remaining diagnostics are lower-level checks behind the headline tables. They are separated from the main results because they support interpretation but are not themselves the paper's primary claims.

The retrieval-proxy table reports deterministic string and URL checks from traces. These proxies help show why answer visibility is not the same as correct evidence use.

\begin{center}
\footnotesize
\setlength{\tabcolsep}{2pt}
\begin{tabularx}{\linewidth}{@{}L{0.45\linewidth}Y@{}}
\toprule
\textbf{Metric} & \textbf{B/T/F} \\
\midrule
Gold URL exact hit & \BraveGoldURLExact{} / \TavilyGoldURLExact{} / \FirecrawlGoldURLExact{} \\
Gold domain hit & \BraveGoldDomain{} / \TavilyGoldDomain{} / \FirecrawlGoldDomain{} \\
Gold source-family hit & \BraveGoldFamily{} / \TavilyGoldFamily{} / \FirecrawlGoldFamily{} \\
Answer in snippet surface & \BraveSnippetSurfaceHit{} / \TavilySnippetSurfaceHit{} / \FirecrawlSnippetSurfaceHit{} \\
Answer in fetched page & \BraveAnswerPage{} / \TavilyAnswerPage{} / \FirecrawlAnswerPage{} \\
Answer in any retrieved text & \BraveAnswerAvailable{} / \TavilyAnswerAvailable{} / \FirecrawlAnswerAvailable{} \\
Wrong despite answer text & \BraveWrongWithAnswer{} / \TavilyWrongWithAnswer{} / \FirecrawlWrongWithAnswer{} \\
\bottomrule
\end{tabularx}
\captionof{table}{Trace-derived retrieval proxies. B/T/F = Brave/Tavily/Firecrawl; these deterministic URL and answer-string checks are not oracle support labels.}
\label{tab:app-retrieval-diagnostics}
\end{center}

The Wilson-interval table expands the decision partition by provider and cell. It is included because some cells are small, so raw cell accuracies should be read with uncertainty.

\begin{center}
\begin{minipage}{\linewidth}
\footnotesize
\begin{tabularx}{\linewidth}{@{}L{0.34\linewidth}Z@{}}
\toprule
\textbf{Provider/cell} & \textbf{Correct rate and Wilson 95\% CI} \\
\midrule
Brave \smart{} & \BraveSmartCorrect{}/\BraveSmartN{}: \BraveSmartRate{} \BraveSmartCI{} \\
Brave \missed{} & \BraveMissedCorrect{}/\BraveMissedN{}: \BraveMissedRate{} \BraveMissedCI{} \\
Brave \blind{} & \BraveBlindCorrect{}/\BraveBlindN{}: \BraveBlindRate{} \BraveBlindCI{} \\
Brave \noop{} & \BraveNoopCorrect{}/\BraveNoopN{}: \BraveNoopRate{} \BraveNoopCI{} \\
Tavily \smart{} & \TavilySmartCorrect{}/\TavilySmartN{}: \TavilySmartRate{} \TavilySmartCI{} \\
Tavily \missed{} & \TavilyMissedCorrect{}/\TavilyMissedN{}: \TavilyMissedRate{} \TavilyMissedCI{} \\
Tavily \blind{} & \TavilyBlindCorrect{}/\TavilyBlindN{}: \TavilyBlindRate{} \TavilyBlindCI{} \\
Tavily \noop{} & \TavilyNoopCorrect{}/\TavilyNoopN{}: \TavilyNoopRate{} \TavilyNoopCI{} \\
Firecrawl \smart{} & \FirecrawlSmartCorrect{}/\FirecrawlSmartN{}: \FirecrawlSmartRate{} \FirecrawlSmartCI{} \\
Firecrawl \missed{} & \FirecrawlMissedCorrect{}/\FirecrawlMissedN{}: \FirecrawlMissedRate{} \FirecrawlMissedCI{} \\
Firecrawl \blind{} & \FirecrawlBlindCorrect{}/\FirecrawlBlindN{}: \FirecrawlBlindRate{} \FirecrawlBlindCI{} \\
Firecrawl \noop{} & \FirecrawlNoopCorrect{}/\FirecrawlNoopN{}: \FirecrawlNoopRate{} \FirecrawlNoopCI{} \\
\bottomrule
\end{tabularx}
\captionof{table}{Decision-cell Wilson 95\% intervals.}
\label{tab:app-wilson}
\end{minipage}
\end{center}

The overlap table gives a compact view of which providers answer the same questions correctly. It supports the complementarity claim by separating shared successes from provider-specific successes.

\begin{center}
\footnotesize
\setlength{\tabcolsep}{2pt}
\begin{tabularx}{\linewidth}{@{}L{0.45\linewidth}Y@{}}
\toprule
\textbf{Pair} & \textbf{both / left / right / wrong} \\
\midrule
Brave vs. Firecrawl & 14 / 11 / 12 / 63 \\
Brave vs. Tavily & 14 / 11 / 11 / 64 \\
Firecrawl vs. Tavily & 14 / 12 / 11 / 63 \\
\bottomrule
\end{tabularx}
\captionof{table}{Semantic-correct pairwise overlap: both / left / right / wrong.}
\label{tab:app-pairwise}
\end{center}

The final diagnostic table reports legacy token F1 and token and fetch volume. These counts ground the paper's cost and retrieval-budget discussion in the actual traced runs.

\begin{center}
\footnotesize
\setlength{\tabcolsep}{2pt}
\begin{tabularx}{\linewidth}{@{}L{0.45\linewidth}Y@{}}
\toprule
\textbf{Metric} & \textbf{B/T/F} \\
\midrule
Token F1 & \BraveFone{} / \TavilyFone{} / \FirecrawlFone{} \\
Total tokens & \BraveTokensM{}M / \TavilyTokensM{}M / \FirecrawlTokensM{}M \\
Median/query & \BraveMedianTokens{} / \TavilyMedianTokens{} / \FirecrawlMedianTokens{} \\
Max/query & \BraveMaxTokens{} / \TavilyMaxTokens{} / \FirecrawlMaxTokens{} \\
$>$100k queries & \BraveOverHundredK{} / \TavilyOverHundredK{} / \FirecrawlOverHundredK{} \\
Fetch success/fail & \BraveFetchSuccess{}/\BraveFetchFailed{} / \TavilyFetchSuccess{}/\TavilyFetchFailed{} / \FirecrawlFetchSuccess{}/\FirecrawlFetchFailed{} \\
\bottomrule
\end{tabularx}
\captionof{table}{Legacy token F1, token usage, and fetch diagnostics. B/T/F = Brave/Tavily/Firecrawl.}
\label{tab:app-token}
\end{center}

\begin{description}[leftmargin=0pt,itemsep=0.12em,topsep=0.1em]
\item[Analysis inputs.] The reported analyses used the semantic-answer audit, three provider trace files, three Kimi per-URL judgment files, and provider-comparison summaries.
\item[Released artifacts.] The release includes code, prompt templates, configurations, sampling procedures, schemas, evaluation and visualization scripts, paper sources, and derived results without copied provider content. Complete traces and content-bearing judge, audit, and trajectory-view records are excluded.
\end{description}